\documentclass[runningheads]{llncs}
\usepackage{graphicx}
\usepackage{amsmath,amssymb} 
\usepackage{color}
\newcommand{\etal}{\textit{et al}. }
\newcommand{\ie}{\textit{i}.\textit{e}., }
\newcommand{\eg}{\textit{e}.\textit{g}., }
\usepackage[width=122mm,left=12mm,paperwidth=146mm,height=193mm,top=12mm,paperheight=217mm]{geometry}
\usepackage{amssymb}
\usepackage{pifont}
\newcommand{\cmark}{\ding{51}}%
\newcommand{\xmark}{\ding{55}}%
\newcommand{\II}{\mathcal{I}}
\newcommand{\DD}{\mathcal{D}}
\newcommand{\PP}{\mathcal{P}}
\newcommand{\OO}{\mathcal{O}}

\usepackage{subcaption}     
\usepackage[utf8]{inputenc} 
\usepackage[T1]{fontenc}    
\usepackage{hyperref}       
\usepackage{url}            
\usepackage{booktabs}       
\usepackage{amsfonts}       
\usepackage{nicefrac}       
\usepackage{microtype}      
\usepackage{epsfig}
\usepackage{graphicx}
\hypersetup{colorlinks=true, linkcolor=red}
\usepackage{wrapfig}
\usepackage{algorithm}
\usepackage{algorithmic}
\usepackage[sort&compress,square,comma,numbers]{natbib}
\makeatletter
\renewcommand\bibsection%
{
  \section*{\refname
    \@mkboth{\MakeUppercase{\refname}}{\MakeUppercase{\refname}}}
}
\makeatother
\usepackage{array}
\newcolumntype{L}[1]{>{\raggedright\let\newline\\\arraybackslash\hspace{0pt}}m{#1}}
\newcolumntype{C}[1]{>{\centering\let\newline\\\arraybackslash\hspace{0pt}}m{#1}}
\newcolumntype{R}[1]{>{\raggedleft\let\newline\\\arraybackslash\hspace{0pt}}m{#1}}

\begin{document}
\pagestyle{headings}
\mainmatter

\def\ACCV18SubNumber{392}  

\title{AVID: Adversarial Visual Irregularity Detection} 
\titlerunning{AVID: Adversarial Visual Irregularity Detection}
\authorrunning{Sabokrou et al.}

\author{Mohammad Sabokrou$^1$, Masoud Pourreza$^2$, Mohsen Fayyaz$^3$, Rahim Entezari$^4$, Mahmood Fathy$^{1}$, Jürgen Gall$^3$, Ehsan Adeli$^5$}
\institute{$^1$Institute for Research in Fundamental Sciences (IPM)\quad$^2$AI \& ML Center of Part\quad$^3$University of Bonn\quad$^4$Complexity Science Hub, Vienna\quad $^5$Stanford University}
\maketitle

\begin{abstract}
Real-time detection of irregularities in visual data is very invaluable and useful in many prospective applications including surveillance, patient monitoring systems, {\it etc}. With the surge of deep learning methods in the recent years, researchers have tried a wide spectrum of methods for different applications. However, for the case of irregularity or anomaly detection in videos, training an end-to-end model is still an open challenge, since often irregularity is not well-defined and there are not enough irregular samples to use during training. In this paper, inspired by the success of generative adversarial networks (GANs) for training deep models in unsupervised or self-supervised settings, we propose an end-to-end deep network for \textit{detection} and \textit{fine localization} of irregularities in videos (and images). Our proposed architecture is composed of two networks, which are trained in competing with each other while collaborating to find the irregularity. One network works as a pixel-level irregularity $\II$npainter, and the other works as a patch-level $\DD$etector. After an adversarial self-supervised training, in which $\II$ tries to fool $\DD$ into accepting its inpainted output as regular (normal), the two networks collaborate to detect and fine-segment the irregularity in any given testing video. Our results on three different datasets show that our method can outperform the state-of-the-art and fine-segment the irregularity. 
\end{abstract}

\section{Introduction}
In the recent years, intelligent surveillance cameras are very much exploited for different applications related to the safety and protection of environments. These cameras are located in sensitive locations to encounter dangerous, forbidden or strange events. Every moment vast amounts of videos are captured by these cameras, almost all of which comprise normal every-day events, and only a tiny portion might be irregular events or behaviors. Accurate and fast detection of such irregular events is very critical in designing a reliable intelligent surveillance system. Almost in all applications, there is no clear definition of what the irregularity can be. The only known piece is whatever that deviates from the normal every-day activities and events in the area should be considered irregularity \cite{boiman2007detecting}. This is a subjective definition that can include a wide-range of diverse events as irregularity and hence makes it hard for automated systems to decide if an event in the scene is really irregularity. Therefore, systems are generally trained to learn the regularity, and rigorously tag everything else as irregularity \cite{mahadevan2010anomaly}. 

Several different methods are used in the literature for learning the normal concept in visual scenes. Low-level visual features such as histogram of oriented gradients (HOG) \cite{bertini2012multi} and histogram of optical flow (HOF) \cite{colque2017histograms,xia2015learning} were the first feature subsets explored for representing regular scenes in videos. Besides, trajectory features \cite{morris2011trajectory} are also used for representing and modeling the videos, although they are not robust against problems like occlusion \cite{sabokrou2016video,bertini2012multi}. Both low-level and trajectory features achieved good results while imposing a relatively high complexity to the system. Recently, with the surge of deep learning methods, several methods are proposed for detecting and localizing irregular events in videos \cite{xia2015learning,you2017provable,sabokrou2017deep,sabokrou2016video,sabokrou2016deep}. 
\begin{figure}[t]
\begin{center}
\includegraphics[width=0.9\linewidth]{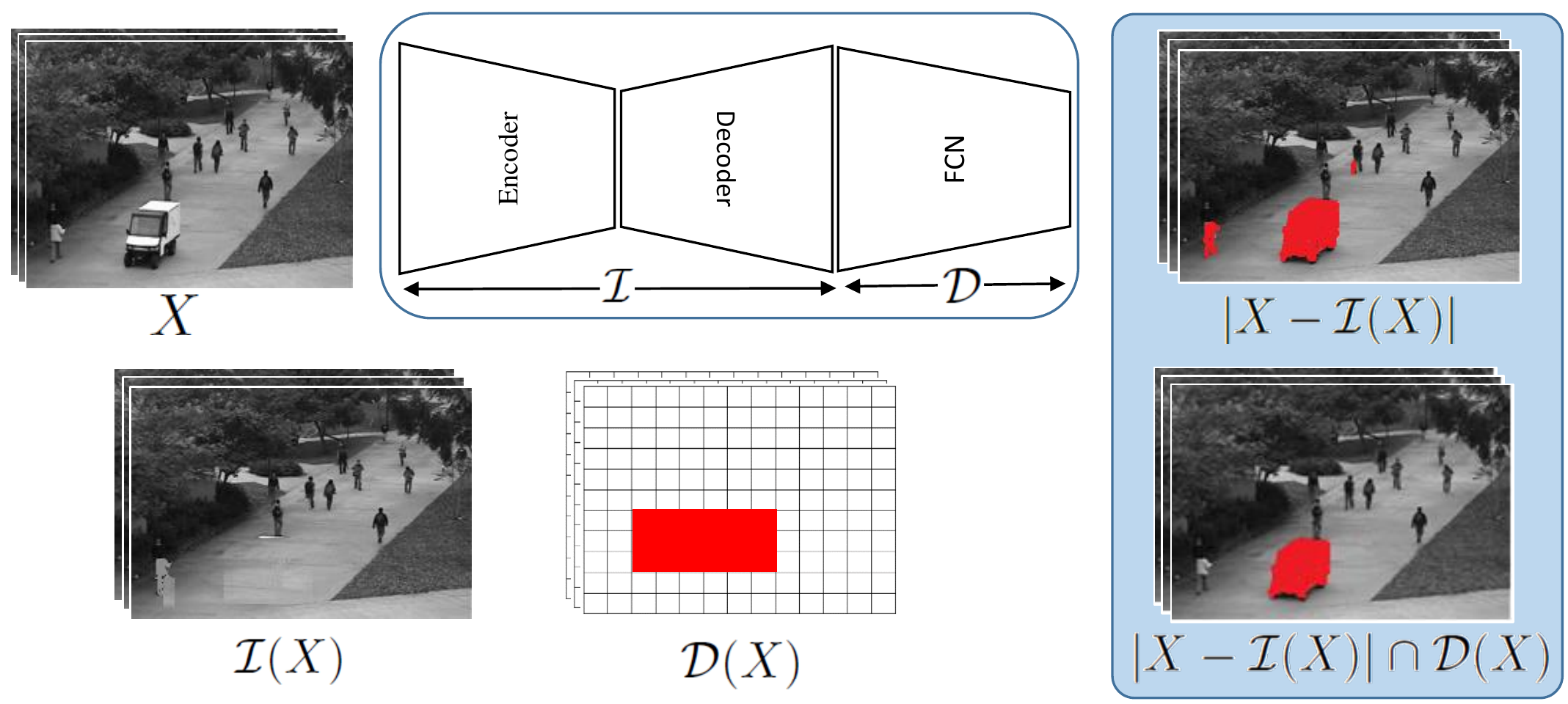}
\end{center}
\vspace{-4mm}
 \caption{The two networks $\II$ and $\DD$ are trained jointly in an adversarial manner. $\II$ is an encoder-decoder convolutional network, which is trained to inpaint its input, $X$, \ie remove the irregularity. Therefore, $|X-\II(X)|$ indicates the pixel-level segmentation of the irregularity, from $\II$'s point-of-view. Whereas, $\DD$ is a fully convolutional network (FCN), which identifies if different regions of its input are normal or irregular (patch-level). The intersection of the pixels denoted as irregularity in both $\II$ and $\DD$ 
 are labeled as the fine-segmentation of irregularity.}
\label{fig:HLS}
\end{figure}

Although these deep learning-based methods effectively advanced the field, they fell short of learning end-to-end models for both detecting the irregularities and localizing them in spatio-temporal sequences, due to several challenges: (1) In applications like irregularity detection, there are little or no training data from the positive class (\ie irregularity), due to the nature of the application. Hence, training supervised models, such as convolutional neural networks (CNNs), is nearly impossible. Therefore, researchers (\eg in \cite{sabokrou2016deep}) have usually utilized pre-trained networks to extract features from the scenes, and the decision is surrendered to another module. (2) To train end-to-end models for this task, just recently \cite{lawson2017finding,sabokrou2018adversarially,schlegl2017unsupervised,ravanbakhsh2017training} used generative adversarial networks (GANs) and adopted unsupervised methods learning the positive class (\ie irregular events). In these methods, two networks (\ie generator and discriminator) are trained. Generator generates data to compensate for the scarcity of the positive class, while the discriminator learns to make the final decision making on whether its input is a normal scene or irregularity. Although they are trained with a very high computational load, the trained generator (or discriminator) is discarded at the testing time. Besides, most of these previous methods are patch-based approaches, and hence are often very slow. Note that these end-to-end models can only classify the scenes and do not precisely localize the irregularities. (3) Accurate pixel-level spatio-temporal localization of irregularities is still an ongoing challenge \cite{sabokrou2017deep}.

In addition to the above issues, as a general and ongoing challenging issue in video irregularity detection, detecting and localizing the irregularity in a pixel-level setting leads to models with high true positives while usually suffering from high false positive errors. On the contrary, some other methods operate on large patches (\eg \cite{bertini2012multi}) and overcome the problem of high false positive error, with the price of sacrificing the detection rate. This motivated us to design a method that takes advantage from both pixel-level and patch-level settings, and come up with a model with high true positive while not sacrificing the detection rate. We do this by proposing an architecture, composed of two networks that are trained in an adversarial manner, the first of which is a pixel-level model and is trained to $\II$npaint its input by removing the irregularity it detects. The second network is a patch-level detector that $\DD$etects irregularities in a patch level. The final irregularity detection and fine-segmentation is, then, defined as a simple intersection of the results from these two networks, having the benefits of both while discarding the pixels that result in high false positive errors (see Fig. \ref{fig:HLS}).

According to the discussions above, in this paper, we propose an end-to-end method for joint detection and localization of irregularities in the videos, denoted as \textit{AVID (Adversarial Visual Irregularity Detection)}. We use an adversarial training scheme, similar to those used in generative adversarial networks (GANs) \cite{goodfellow2014generative}. But in contrast to previous GAN-based models (\eg \cite{lawson2017finding,ravanbakhsh2017training,ravanbakhsh2017abnormal,schlegl2017unsupervised}), we show how the two networks ($\II$ and $\DD$) can help each other to conduct the ultimate task of visual irregularity detection and localization. The two networks can be efficiently learned against each other, where $\II$ tries to inpaint the image such that $\DD$ does not detect the whole generated image as irregularity. By regulating each other, these two networks are trained in a self-supervised manner \cite{odena2016semi,do2018self,sabokrou2018adversarially}. Although, $\II$ and $\DD$ compete with each other during training, they are trained to detect the video irregularity from different aspects. Hence, during testing, these two networks collaborate in detection and fine-segmentation of the irregularity.

In summary, the main contributions of this paper are three-fold: (1) We propose an end-to-end deep network for detection and localization of irregularities in visual data. To the best of our knowledge, this is the first work that operates on a video frame as a whole in and end-to-end manner (not on a patch level). (2) Our method can accurately localize the fine segments of the video that contain the irregularity. (3) Our proposed adversarial training of the two networks (one pixel-level and one patch-level) alleviates the high false positive rates of pixel-level methods while not suffering from high detection error rate of patch-level models.
\vspace{-3mm}
\section{Related Works}
Detection of visual irregularities is closely related to different methods in the literatures (including one-class classifiers, anomaly detection, outlier detection or removal methods). These approaches all search for an irregularity, which is hardly and scarcely seen in the data.  Traditional methods often learn a model for the normal class, and reject everything else (\ie, identify as irregularity). Learning under a constraint (such as sparsity and compressed sensing) or statistical modeling are two common methods for modeling the normal class. For the case of visual data, feature representation (from videos and images) is an important part. Low-level features (such as HOG and HOF) and high-level ones (\eg trajectory) are widely used in the literature. In the recent years, similar to other computer vision tasks, deeply learned features are vastly utilized for irregularity detection. In this section, a brief review of the state-of-the-art methods for irregularity detection and related fields is provided. 

\noindent\textbf{Video Representation for Irregularity Detection.} As one of the earliest representations for irregularity detection, trajectory were used \cite{piciarelli2006line,morris2011trajectory}, such that an event not following a learned normal trajectory pattern is considerd as anomaly. Optical-flows  \cite{colque2017histograms,adam2008robust,cong2013video,benezeth2009abnormal}, social forces (SF) \cite{mehran2009abnormal}, gradient features \cite{bertini2012multi,kratz2009anomaly}, mixture of dynamic textures \cite{mahadevan2010anomaly}, and mixture of probabilistic PCAs (MPPCA) \cite{kim2009observe} are types of low-level motion representations used to model regular concepts. Deep learned features, using auto-encoders \cite{xu2015learning,sabokrou2015real} pre-trained networks \cite{sabokrou2017deep}, or PCAnet  \cite{feng2017learning,fang2016abnormal} have recently shown great success for anomaly detection.

\noindent\textbf{Constrained reconstruction as supervision.} Representation learning for the normal (\ie regular) class under a constraint has shown effective to detect irregular events in visual data. If the new testing data does not conform to the constraint, it can potentially be considered as an irregularity. Learning to reconstruct normal concept with spare representation (\eg in \cite{cong2011sparse}) and minimum effort (\eg in \cite{boiman2007detecting}) are widely exploited for this task. Boiman and Irani \cite{boiman2007detecting} consider an event as irregular if its reconstruction using the previous observations is nearly impossible. In \cite{antic2011video}, a scene parsing approach is proposed by Antic \etal, in which all object hypotheses for the foreground of a frame are explained by normal training. Those hypotheses that cannot be explained by normal training are considered as anomaly. In \cite{sabokrou2016video,sabokrou2018adversarially,cong2011sparse} normal class is learned through a model by reconstructing samples with minimum reconstruction errors. High reconstruction error for a testing sample means this samples is irregular. Also, \cite{sabokrou2016video,cong2011sparse} introduced a  self-representation techniques for video anomaly and outlier detection through sparse representation, as a measure for separating inlier and outlier samples.

\noindent\textbf{Deep Adversarial Learning.} Recently, GANs \cite{goodfellow2014generative} are widely being used for generating data to learn specific models. They are extended for prediction tasks, in which there are not enough data present for training  (\eg, in \cite{lawson2017finding,schlegl2017unsupervised,ravanbakhsh2017training}). GANs are based on a game between two different networks, one generator ($G$) and one discriminator ($D$). $G$ aims to generate sensible data, while $D$ tries to discriminate real data from the fake data generated by $G$. A closely related type of GANs to our work is the conditional GANs \cite{mirza2014conditional}. In conditional GANs, $G$ takes an image $X$ as the input and generates a new image $X'$, whereas, $D$ tries to distinguish $X$ from $X'$. Isola \etal~\cite{isola2016image} proposed an `Image-to-image translation' framework based on conditional GANs, where both $G$ and $D$ are conditioned on the real data. Using a U-Net encoder-decoder \cite{ronneberger2015u} as the generator and a patch-based discriminator, they transformed images with respect to different representations. In another work, \cite{ravanbakhsh2017abnormal} proposed to learn the generator as the reconstructor for normal events, and tag chucks of the input frame as anomaly if they cannot be properly reconstructed. In our work, $\II$ learns to inpaint its input and make it free from irregularity in pixel-level, and $\DD$ regulates it by checking if its output is irregularity or not. This self-supervised learning scheme leads to two networks that improve the detection and fine-segmentation performance for any given testing image.  Liu \etal~\cite{liu2017future} proposed to learn an encoder-decoder GAN to generate the future video frame using optical-flow features, used for irregularity detection, \ie, if the prediction is far from the real future frame, it is counted as irregularity. Similar to all other works, the work in \cite{liu2017future} ignores the discriminator in the testing phase. Also they suffer from high false positive rates. 

\section{AVID: Adversarial Visual Irregularity Detection}
The proposed method for irregularity detection and localization is composed of two  main components: $\II$ and $\DD$. $\II$ learns to remove the pixel-wise irregularity from its input frame (\ie $\II$npaint the video), while $\DD$ predicts the likelihood of different regions of the video (patches) being an irregularity. These networks are learned in an adversarial and self-supervised method in an end-to-end setting. In the following, we outline the details of each network. An overall sketch of the proposed method is illustrated in Fig. \ref{fig:HLS}. In summary, $\II$ learns to $\II$npaint its input $X$ to fool $\DD$ that the inpainted version does not have any irregularities. For $\II$ to learn to reconstruct skewed images, $\II$ is exposed to noisy versions of the videos in the data set and therefore it implicitly learns not only to remove the irregularity but also to remove the noise in the data. Besides, $\DD$ knows the distribution of original data $\PP_d$, as it has access to the data set containing all normal videos (or with a tiny portion of irregularities present in the data). Having access to $\PP_d$, $\DD$ simply rejects poorly inpainted or reconstructed data. These two networks self-supervise each other and are trained through this bilateral game. This structure is inspired by GAN models, although our model does not generate from scratch and only enhances its input tailored for detection of irregularities. 

After the adversarial training, $\II$ will be an expert to inpaint its input (and make it devoid of noise), which successfully fools $\DD$. Module $\II$ is a pixel-level irregularity detector and $\DD$ a patch-level one, hence, $|X - \II(X)| \cap \DD(X)$ can define the fine-segmentation and the precise location of the irregularity in any input testing video frame $X$. Note that each of the two networks $\II$ and $\DD$ can be exploited for detecting and localizing the irregularity, but by aggregating them, we show a great improvement in the results. Detailed descriptions of each module along with the training and testing procedures are explained in the following. 

\subsection{$\II$: Inpainting Network}
In some recent previous works \cite{xia2015learning,sabokrou2016video,sabokrou2018adversarially}, it is stated that when an auto-encoder is trained only on the inlier or normal class, the auto-encoder will be unable to reconstruct outlier or anomaly samples. Since parameters of auto-encoder are optimized to reconstruct samples from the normal (regular) class, as a side-effect, the reconstruction error of outliers (irregularities in our case) will be high. Also, in \cite{sabokrou2018adversarially} in an unsupervised GAN-style training, a patch-based CNN is proposed that decimates outliers while enhancing the inlier samples. This makes the separation between the inliers and outliers much easier. In this paper, we use a similar idea, but in contrast: (1) $\II$ (analogous to the generator in GANs) is not directly used as an irregularity detector; (2) Instead of decimating outliers (irregularities in our case), our network inpaints its input by removing the irregularity from it. Implicitly, $\II$ operates similar to a de-noising network, which replaces the irregularity in the video with a dominant concept (\eg dominant textures).

Architecturally, $\II$ is an encoder-decoder convolutional network (implemented identical to U-Net \cite{ronneberger2015u}), which is trained only with data from the normal (regular) class. It learns to map any given input to the normal concept. Usually, irregularity occurs in some parts of video frames, and $\II$ acts by reconstructing those deteriorated parts of images and videos. 

\begin{figure}[t]
\begin{center}
\includegraphics[width=\linewidth]{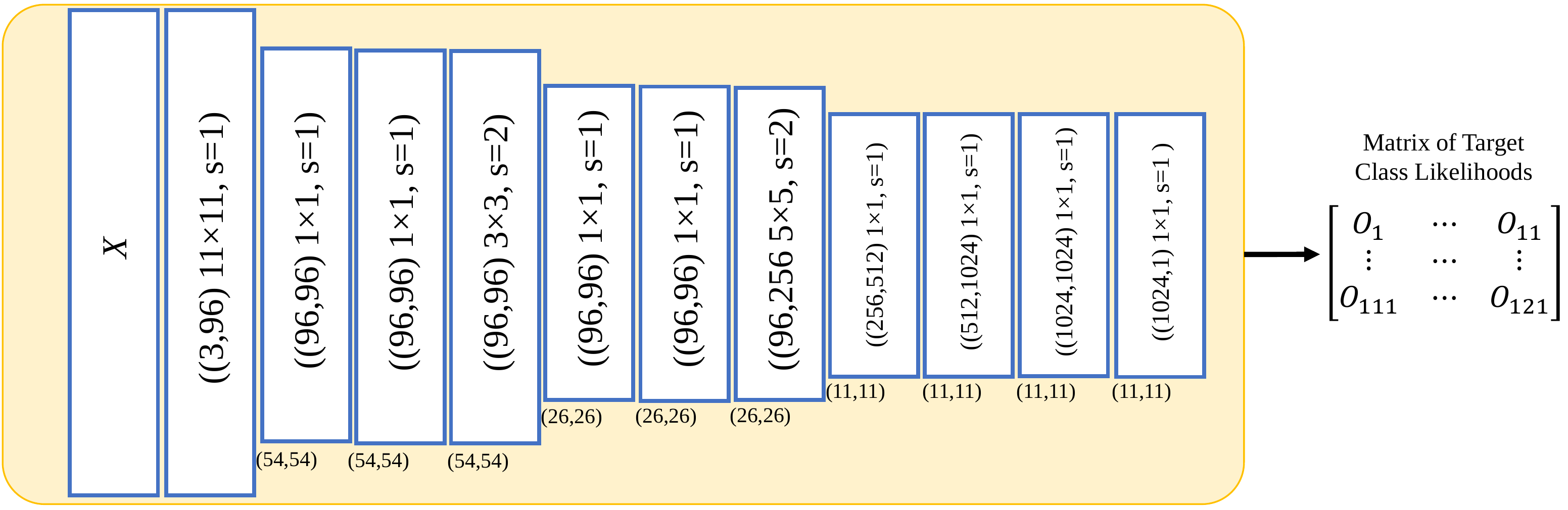}
\end{center}\vspace{-4mm}
 \caption{Structure of $\DD$, a FCN that scores video regions on how likely they are irregularities. All layers are convolutional layers and are described in this form $((C_1,C_2),K, s)$, with $C_1$ and $C_2$ as the number of inputs and outputs of the channel, $K$ size of the applied kernel, and $s$ as the kernels with which stride are convoluted on its input. Underneath each layer, the size of the feature maps are provided. Matrix $\mathcal{O}$, output of $\DD$, defines regularity likelihood for each region.}
\label{fig:d_arch}
\end{figure}

\subsection{$\DD$: Detection Network}
Fully convolutional neural networks (FCNs) can effectively represent video frames and patches, and are previously used for different applications, such as semantic segmentation \cite{long2015fully,nie20183}. In a recent work, \cite{sabokrou2016deep} used a FCN for irregularity detection in videos, in which the authors used a pre-trained FCN just for describing the video patches. Thier method was not capable to detect (or score) the irregularity in the frames. Inspired by this idea, we use a FCN for the detection phase, but train it in an adversarial manner (along with $\II$). Our model is, hence, an end-to-end trainable network. We train the FCN (\ie $\DD$ network) to score (and hence detect) all irregular regions in the input frame all at once.

Unlike conventional discriminator networks in a GAN, where  the discriminator just provides a judgment about its input as a whole, $\DD$ is capable to judge about different regions of its input. Consequently, its output is a matrix of likelihoods, which imply if the regions of its input follow the distribution of the normal (regular) data or not (\ie $\PP_d$). Fig. \ref{fig:d_arch} shows the architecture of $\DD$, which includes several convolutional layers. For this application, since local spatial characteristics are crucial, we do not use any pooling or fully connected layers, which ignore the spatial characteristics. On the other hand, to preserve the locality and enrich the features, several 1$\times$1  convolutional layers are used.

\subsection{Adversarial Training of $\II$+$\DD$ }
Goodfellow \etal~\cite{goodfellow2014generative} proposed an efficient way for training two deep-neural networks (Generator ($G$) and Discriminator ($D$), in their terminology) through adversarial training, called GAN. GANs aim to learn the distribution of training data $\PP_d$, and simultaneously generate new samples based on the same distribution $\PP_d$. Therefore, $G$ maps a vector of random variables (say $Z$) from a specific distribution $\PP_Z$ to a sample from real data distribution and $D$ seeks to discriminate between the actual data and the fake data generated by $G$. Generator and Discriminator are learned in a two-player mini-max game, formulated as:
\begin{equation}
\begin{aligned}
\min_{G} \max_{D} ~& \Big( \mathbb{E}_{X \sim  \PP_d}[\log(D(X))] + \mathbb{E}_{Z \sim  \PP_{z}}[\log(1-D(G(Z)))] \Big).
\end{aligned}
\end{equation}

Similarly, $\II$+$\DD$ can be adversarially trained. Unlike conventional GAN, instead of mapping a latent space $Z$ to a sample from $\PP_d$, $\II$ maps very noisy sample $X+\eta$ to a noisy-less one that can fool $\DD$ into identifying it as a normal scene \ie
\begin{equation}
\tilde{X} = \left(X \sim  \PP_d \right) +\gamma \left(\eta \sim  \mathcal{N}(0, \sigma^2\mathbf{I}) \right) \longrightarrow X' \sim  \PP_d,
\end{equation}
where $\eta$ is a Gaussian noise sampled from the normal distribution with standard deviation $\sigma$, \ie $\mathcal{N}(0, \sigma^2\mathbf{I})$. $\gamma$ is a hyperparameter that defines how severely to contaminate $X$ with noise. Note that the addition of $\eta$ forces $\II$ to learn how to restore $X$ from $\tilde{X}$, \ie in absent of irregularity.

On the other hand, $\DD$ has access to the original pool of training data, hence, knows $\PP_d$, and is trained to identify the normal class. In our case, $\DD$ decides which region of $\II(\tilde{X})$ follows from $\PP_d$. To fool $\DD$, $\II$ is implicitly forced to inpaint its input. As mentioned above, $\DD$ (\ie our discriminator network) judges on the regions of its input and not the whole image (which is the case for the GAN discriminators). Consequently, output of $\DD(X)$ is a matrix, $\OO \in \mathbb{R}^{n_1\times n_2}$, in which each cell $\OO(i,j)$ corresponds to the $i^\text{th}$ and $j^\text{th}$ image region. Therefore, the joint training aims to maximize $\sum_{i=1,j=1}^{i=n_1,j=n_2}\OO(i,j)$ (\ie maximize the likelihood of $\II$'s output to be normal). $n_1\times n_2=n$ is the total number of regions judged by $\DD$. Accordingly, $\II$+$\DD$ can be jointly learned by optimizing the following objective:
\begin{equation}
\begin{aligned}
\min_{\II} \max_{\DD} ~& \Big( \mathbb{E}_{X \sim  \PP_d}[\log(\|\DD(X)\|^2)]  +  \mathbb{E}_{\tilde{X}\sim  \PP_d+\mathcal{N}_\sigma}[\log(\|\mathcal{Y}-\DD(\II(\tilde{X}))\|^2)] \Big),
\end{aligned}
\end{equation}
where $\mathcal{Y} \in \mathbb{R}^{n_1 \times n_2}:=\mathbf{1}^{n_1 \times n_2}$. Based on the above objective function, $\II$ learns to generate samples with the distribution of normal data (\ie $\PP_d$). Hence, the parameters of this network, $\theta_\II$, are learned to restore a noisy visual sample. So, $\II(\tilde{X},\theta_\II)$ would be an irregularity-less version of $\tilde{X}$. For better understanding, suppose every frames of the video $X$ is partitioned into $n=n_1\times n_2$ non-overlapping regions (blocks), $B_{i\in1:n}$. The proposed algorithm looks to find which of these regions are irregular. After the joint training of $\II$+$\DD$, the modules can be interpreted as follows:

\begin{figure}[t]
\begin{center}
\addtolength{\tabcolsep}{-5pt}    
\begin{tabular}{ccccccc}
    & \multicolumn{3}{c}{} &  & \multicolumn{2}{c}{} \\
{\footnotesize $X$} ~~& 
\raisebox{-.5\height}{\includegraphics[width=0.25\linewidth]{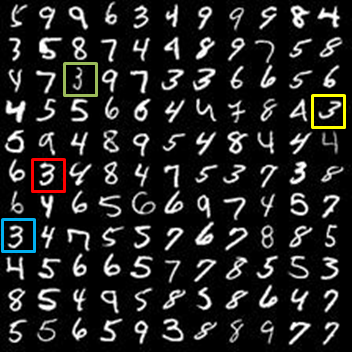}} &~~
\raisebox{-.5\height}{\includegraphics[width=0.25\linewidth]{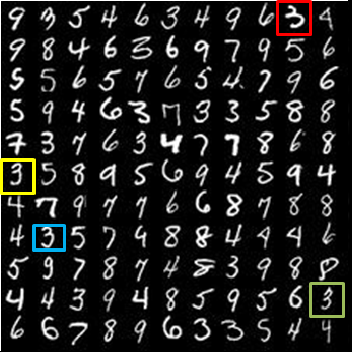}} &~~ \raisebox{-.5\height}{\includegraphics[width=0.25\linewidth]{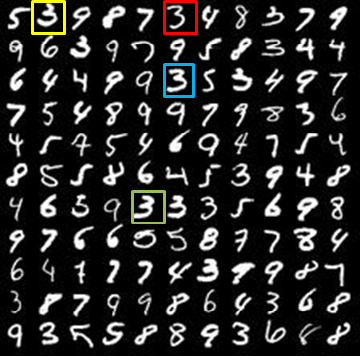}}  \\ \
{\footnotesize $\II({X})$}~~ ~~& \raisebox{-.5\height}{\includegraphics[width=0.25\linewidth]{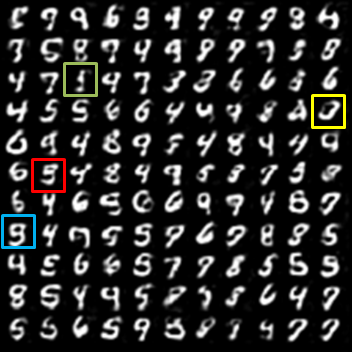}} &~~ \raisebox{-.5\height}{\includegraphics[width=0.25\linewidth]{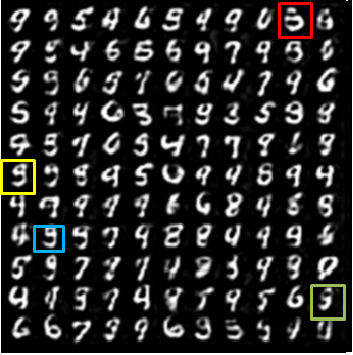}} & ~~ \raisebox{-.5\height}{\includegraphics[width=0.25\linewidth]{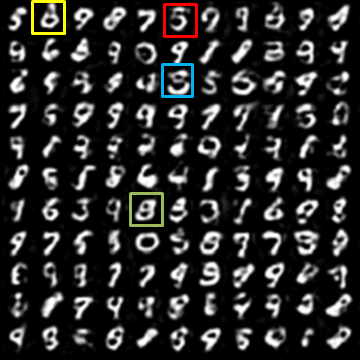}} \\ 
\end{tabular}
\addtolength{\tabcolsep}{5pt}    
\end{center}\vspace{-4mm}
   \caption{Examples of images ($X$) and their inpainted versions using $\II$ (\ie $\II(X)$). The network is trained on images containing 0-9 digits, except digit `3'. These images are created from images in the MNIST dataset \cite{lecun2010mnist}, to show how $\II$ and $\DD$ operate. When digit `3' appears in a test image, it is considered as an irregularity. For clarity, several irregularity regions are marked in $X$ and $\II(X)$.}
 \label{fig:ir-mnist}
\end{figure}

\begin{figure}[t]
\begin{center}
\includegraphics[width=10cm, height=2cm]{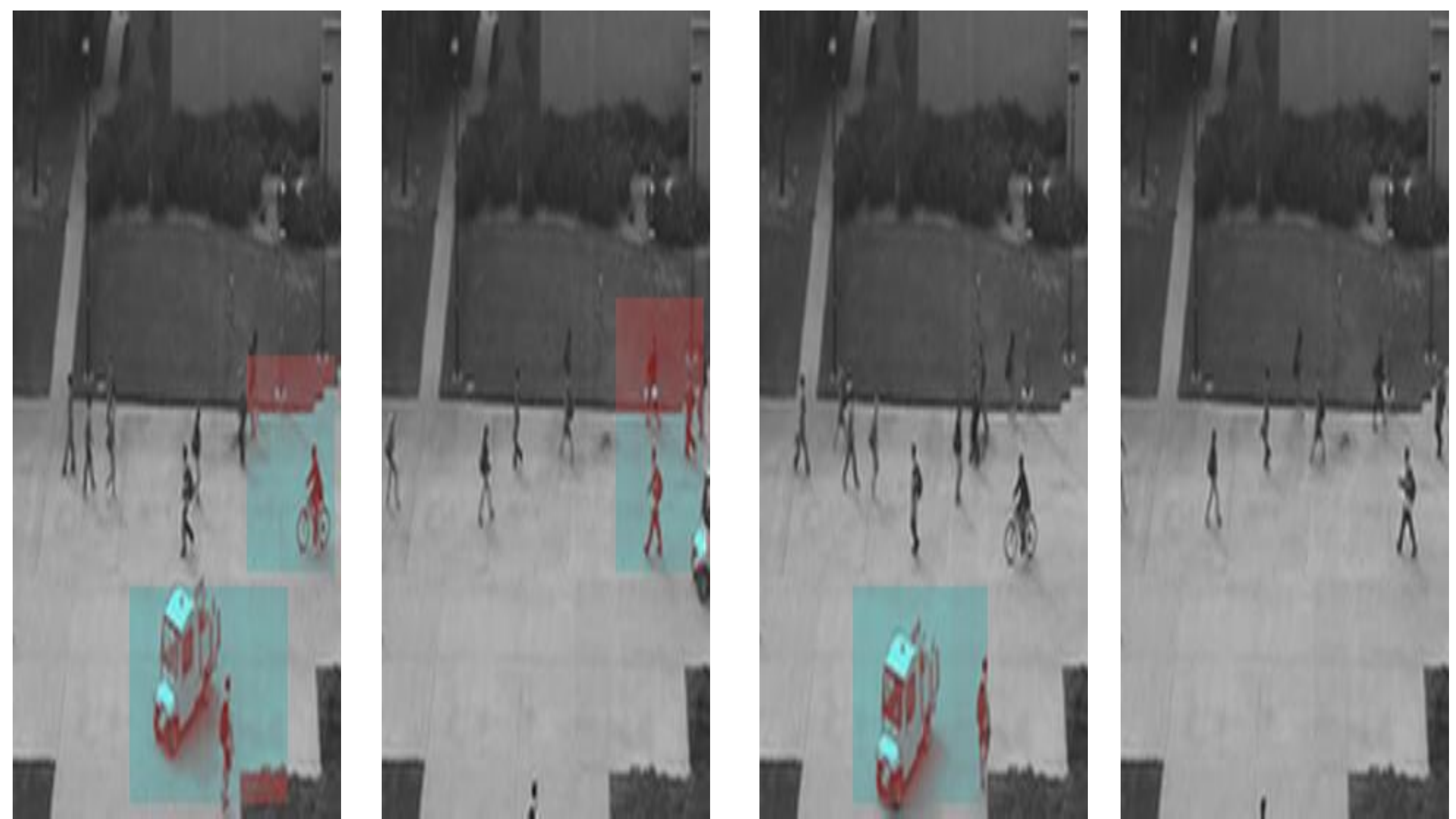}
\end{center}\vspace{-4mm}
 \caption{
 Examples of the output of $\DD$ , \ie matrix $\OO$, mapped on the  original frames. The colored areas on the image are the low-scored regions in $\OO$.
 }
\label{fig:d_z}
\end{figure}

\begin{itemize}
    \item{$\forall{i}$;~$B_i \sim \PP_d+\mathcal{N}_\sigma$; $\II(\tilde{X}=B_{i\in1:n}) \Rightarrow X'=B'_{i\in1:n}$, where $\forall{i}$;~$B'_{i}\sim \PP_d$. This is the case that the input is free from irregularity and is already following $\PP_d$. $X'$ is the output of $\II$, and hence $\|X-X'\|$ is minimized (will be near zero). This is because of the fact that $\theta_{\II}$ is optimized to reconstruct its input (all $B_i$ regions) while the output also follows $\PP_d$. Note that $\II$ works similar but not exactly the same as the refinement network in \cite{sabokrou2018adversarially}, the de-noising auto-encoder in \cite{sabokrou2016video}, or de-nosing convolutional neural network in \cite{divakar2017image}. Consequently, if the input frame is already free from irregularity, $\II$ acts as only an enhancement function.}
    \item{$\exists{j}$;	$B_j \nsim \PP_d$; $\II(\tilde{X}=B_{i\in1:n} ) \Rightarrow X'=B'_{i\in1:n}$, where $\forall i;~B'_{i}\sim \PP_d$. This is the case that at least one of the regions in $\tilde{X}$ is irregularity. Then, it is expected from $\II$ that $\II({B}_{j}\nsim \PP_d) \Rightarrow B'_{j} \sim \PP_d$. Irregularity region is forced to follow from the normal data distribution, as $\II$ is trained to restore a normal region contaminated with a hard noise ($\gamma (\eta \sim  \mathcal{N}(0, \sigma^2\mathbf{I}))$) to a clean noise-less normal-looking region. In the testing phase, an irregularity region, in $\II$'s point-of-view, is consider as a hard noise. Note that in our experiments, $\gamma < 0.4$ is considered as soft and $\gamma \geq 0.4 $ is considered as hard noise. Hard noises added to the training samples (inputs of $\II$) are considered as concepts that should be removed from output of $\II$, for it to be able to fool $\DD$. See Fig. \ref{fig:ir-mnist}, as a proof-of-concept example. Digit `3' is considered as an irregular concept in this example, and $\II$+$\DD$ have never seen any `3' during training. So, $\II$ tries to replace it with a dominant concept from the normal concepts, which can be any digit between 0-9 except `3'. Consequently those $B_i$s that follow $\PP_d$ are not touched (or are enhanced), while those not following from the normal data distribution are converted to a dominant concept (\ie inpainted).}
    \item{$\DD(X=B_{i\in1:n})\Rightarrow \OO_{i\in1:n}$ where each element of matrix $\OO$, output of $\DD$, indicates score for the corresponding region to be normal (regularity). Note that here $\OO_i$ is analogous to $\OO(a,b)$ with $i=b+(a-1)\cdot n_1$. Let's consider $\exists{j}$; $B_j \nsim \PP_d$; we expect that $\OO_j \leq \mathcal{O}_{i \neq j}$. Parameters of $\DD$, $\theta_\DD$, are learned to map normal regions (\ie following $\PP_d$) to 1, and 0 otherwise. Fig. \ref{fig:d_z} shows the results of $\DD$, in which the locations with an irregularity have lower scores.}  
\end{itemize}

With a modification on the objective function and the structure of GANs, our two proposed deep networks are adversarially trained. They learn to identify irregularities from two different aspects (pixel-level and patch-level) in a self-supervised manner. The equilibrium point as the stopping criterion for the two networks is discussed in Section \ref{sec:discussion}.

\subsection{Irregularity Detection}
\label{sec:AID}
In the previous subsections, detailed structures and behaviours of the two networks are explained. As mentioned, $\II$ acts as a pixel-level inpainting network, and $\DD$ as a patch-level irregularity detector.   The difference between the input and outputs of $\II$ for any testing frame $X$ (\ie $|X-\II(X)|$) can be a guideline for where pixels of the input frames are irregular. On the other hand, $\DD(X=B_{i\in1:n})$ shows which regions of $X$ are irregular (\ie those with $\OO_j\leq \zeta$). As discussed earlier, the detection based on $\II$ leads to high false positive, and solely based on $\DD$ leads to high detection error rates. Therefore, outputs of these two networks are masked by each other and the intersection is considered as the irregularity localization.

To identify the regions of irregularities from the output of $\DD$ (\ie matrix $\OO$), we can consider all regions that $\{B_i|(\OO_{i} \leq \zeta)\}$, where $B_i$ is respective field of $\OO_i$ on the input. As mentioned above, $\II$ will reconstruct its whole input image, except for the irregularities, which are inpainted. Consequently, pixels where  $|X-\II(X)| \geq \alpha$ can be considered as potential pixels containing an irregularity. To alleviate the high false positive rate, we just mask these pixels with the above regions. Consequently, final irregularity fine-segmentation on $X$ can be defined as
\begin{equation}
\label{eq:2}
    \mathcal{M}=\{|X-\II(X)|\geq \alpha\} \cap \{B_i|(\OO_{i} \leq \zeta)\}.
\end{equation}

\subsection{Preprocessing of the Videos}
Irregular events in visual data (especially in videos) are defined in terms of irregular shapes, motion, or possibly a combination of both. Therefore, to identify the motion properties of events, we require a series of frames. Two strategies can be adopted for this purpose: (1) Adding a LSTM sequence at the end of current proposed network \cite{sutskever2014sequence}; (2) Using 3D kernels, instead of 2D ones in the same architectures we proposed (such as in \cite{ji20133d}). However, these methods increase the number of parameters and the computational complexity of the model. \cite{sabokrou2016deep} proposed a simple preprocessing step to feed videos instead of images to a CNN without any modification on the structures of a 2D CNN. To interpret both shape and motion, we consider the pixel-wise {\it average} of frame $I_{t}$ and previous frame $I_{t-1}$, denoted by $I^\prime_{t}$ (not to be confused with a derivative): $I^\prime_{t}(p)=\frac{1}{2}({I_{t}(p)+I_{t-1}(p)})$, where $I_{t}$ is the $t^{th}$ frame in the video. For detecting irregularities on $I_{t}$, we use the sequence $X=\langle I'_{t-4},I'_{t-2},I'_{t}\rangle$, and input it to the three channels (similar to R, G, and B channels) of the networks.

\section{Experimental Results}
We evaluate the performance of the proposed method on two standard benchmarks for detecting and localizing irregular event in videos. Also, we create a dataset, called IR-MNIST, for evaluating and analyzing the proposed method to better showcase the abilities of the network modules, as a proof-of-concept. 

The proposed method is implemented using PyTorch \cite{paszke2017automatic} framework. All reported results are from implementations on a GeForce GTX 1080 GPU. Learning rate of $\II$ and $\DD$ set to be the same and is equal to 0.002. Also, momentum of both is equal to 0.9, with a batch size of 16. The hyperparameter $\gamma$, which controls the scale of added Gaussian noise for the training samples (in $\II$) is equal to 0.4. 
\subsection{Datasets}
\noindent\textbf{UCSD:} This dataset \cite{mahadevan2010anomaly} includes two subsets, Ped1 and Ped2 . Videos are from outdoor scene, recorded with a static camera at 10 fps. The dominant mobile objects in these scenes are pedestrians. Therefore, all other objects (\eg cars, skateboarders, wheelchairs, or bicycles) are considered as irregularities.
 
\noindent\textbf{UMN:} UMN dataset is recorded for three different scenarios. In each scenario, a group of people walk in normal pace in an area, but suddenly all people run away (\ie they escape). The escape is considered to be the irregularity. 

\noindent\textbf{IR-MNIST (Available at \url{http://ai.stanford.edu/~eadeli/publications/data/IR-MNIST.zip}):} To show the proprieties of the proposed method, we create a simple dataset using the images from MNIST \cite{lecun1998gradient}. To create each single image, randomly 121 samples are selected from the MNIST dataset and are put together as a $11\times11$ puzzle. Some samples are shown in Fig. \ref{fig:ir-mnist}. We create as much images to have 5000 training data and 1000 test samples. Training samples are created without using any images of the digit `3'. Hence, `3' is considered as an irregular concept. We expect that our method detects and localizes all patches containing `3' in the testing images, as irregularity.
\subsection{Results}

\begin{figure}[t]
\begin{center}
\includegraphics[width=11cm, height=4cm]{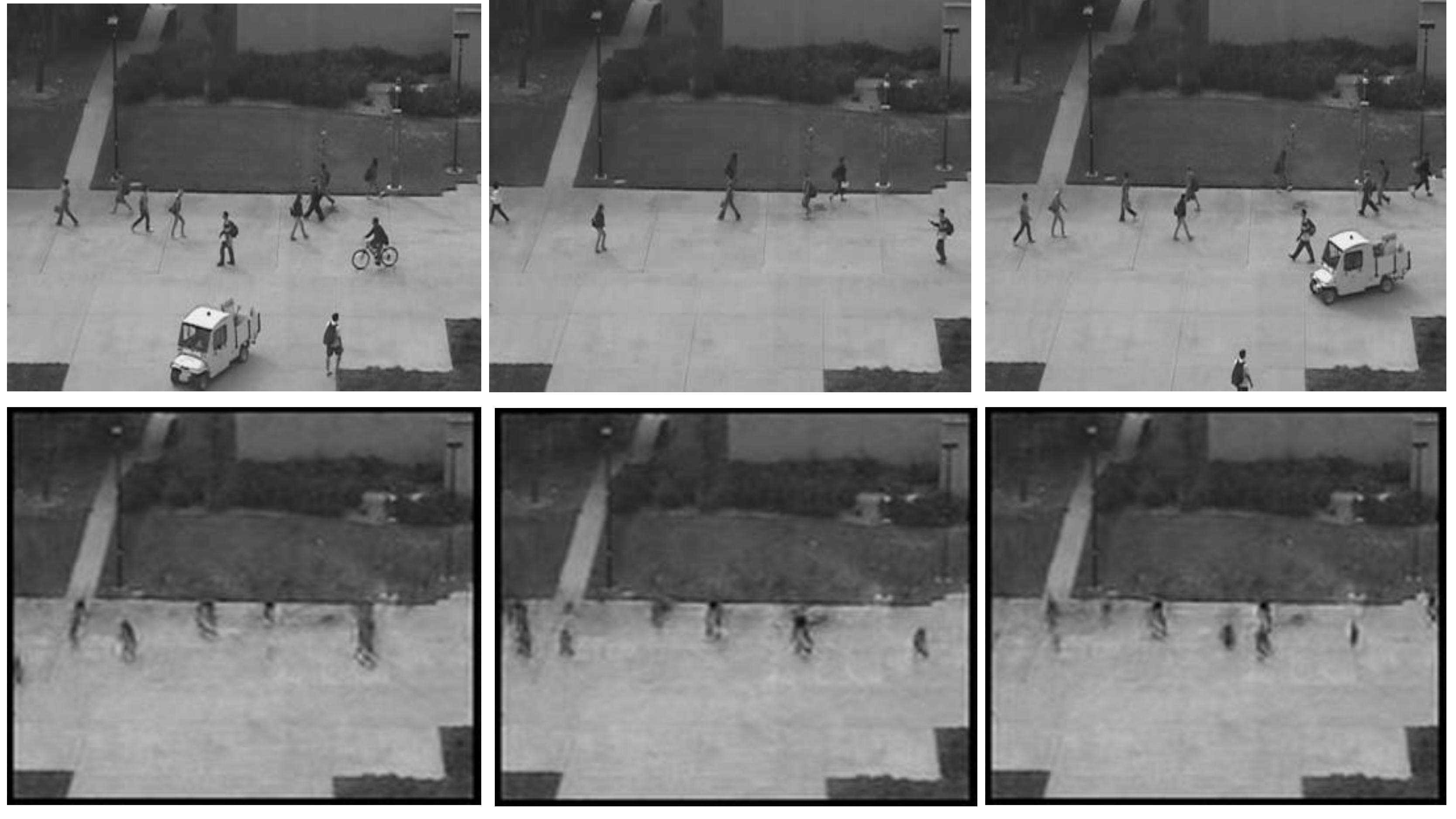}
\end{center}\vspace{-4mm}
 \caption{Examples of the output of $\II$ on UCSD dataset. Bottom row shows output of $\II$. Top row shows the original frames.}
\label{fig:g_ucsd}
\end{figure}

\begin{table}[t]
\caption{Frame-level (FL) and pixel-level (PL) comparisons on UCSD dataset. Last column shows if methods are (1) based on $\mathbb{D}$eep learning or not, (2) $\mathbb{E}$nd-to-end deep network or not, and finally (3)  $\mathbb{P}$atched based method or not.}
\begin{center}\vspace{-3mm}
\begin{tabular}{lccccc}
\hline
Method & Ped1 (FL/PL) & Ped2 (FL/PL)  &  ($\mathbb{D}$/$\mathbb{E}$/$\mathbb{P}$)\\ \hline \hline
IBC\cite{boiman2007detecting} &(14/26) &(13/26) & (\xmark/\xmark/\cmark)\\
MDT\cite{mahadevan2010anomaly}&(25/58)&(24/54)&(\xmark/\xmark/\xmark)\\
Bertini \etal \cite{bertini2012multi} &(31/70) &(30/--) &(\xmark/\xmark/\cmark)\\
Xu \etal \cite{xu2014video} &(22/--)&(20/42) &(\xmark/\xmark/\xmark) \\
Li \etal \cite{li2014anomaly} & (16/--)&(18/29)& (\xmark/\xmark/\xmark)\\ \hline \hline
RE \cite{sabokrou2016video} & (--/--)&(15/--)&(\cmark/\xmark/\cmark)\\
Xu \etal \cite{xu2015learning}&(16/40)&(17/42)&(\cmark/\xmark/\cmark)\\
Sabokrou \etal \cite{sabokrou2015real} &(--/--)&(19/24)&(\cmark/\xmark/\cmark) \\
Deep-Cascade\cite{sabokrou2017deep} &(9.1/15.8) &(8.2/19)&(\cmark/\xmark/\cmark) \\
Deep-Anomaly\cite{sabokrou2016deep} & (--/--)&(\textbf{11}/\textbf{15})&(\cmark/\xmark/\xmark)\\ 
Ravanbakhsh \etal \cite{ravanbakhsh2017training}   &(\textbf{7}/34)& (\textbf{11}/--) & (\cmark/\xmark/\cmark) \\ 
ALOCC \cite{sabokrou2018adversarially}&(--/--)&(13/--)& (\cmark/\cmark/\cmark)\\ \hline \hline
$\DD(X)$&(--/16.7)& (--/17.2 )&(\cmark/\cmark/\xmark)\\
$\II(X)$ &(--/17.3)& (--/17.8)&(\cmark/\cmark/\xmark)\\
AVID &(12.3/\textbf{14.4})&(14/\textbf{15}) &(\cmark/\cmark/\xmark)\\

\end{tabular}
\end{center}

\label{tab:EER}

\end{table}
\noindent\textbf{Results on UCSD.}
Fig. \ref{fig:g_ucsd} visualizes the outputs of network $\II$ on several examples of UCSD frames. As can be seen, irregular objects such as bicycles and cars are disappeared in the output of $\II(X)$, and the regular regions are approximately reconstructed. Note that, however $\II$ is trained to reconstruct regular regions with minimum loss, loss of quality is unavoidable, as a consequence of the hard noise applied to the inputs of $\II$ during training. This shortage, however, does not adversely affect our final decision, because maximum  difference between $X$ and $\II(X)$ happens in the pixels that an irregularity occurred. 
Fig. \ref{fig:d_z} also shows several output samples of the proposed detector $\DD$ for detecting irregularity in videos. It confirms that irregular blocks can be appropriately detected. For quantitative analysis, similar to \cite{mahadevan2010anomaly}, two standard measures on this dataset are considered. In frame level (FL) each of the frames is considered to be anomaly if at least one pixel is detected as irregularity. In pixel-level (PL) analysis, the region identified as anomaly should have an overlap of at least 40\% with the ground-truth irregularity pixels to be considered as irregularity. PL is a measure for evaluating the accuracy of the localization in a pixel-level. A comparison between performance of the proposed and the state-of-the-art methods is provided in Table \ref{tab:EER}.  The proposed method for detecting the irregular frames is comparable to state-of-the-art method, but the localization performance outperforms all other methods by a large margin. As can be seen,  \cite{sabokrou2017deep,ravanbakhsh2017training,sabokrou2018adversarially} achieve a better performance by a narrow margin in a frame-level aspect compared to us, but unlike ours, these methods are not able to process images as a whole in an end-to-end fashion. They require to split a frame into a set of patches and feed them to the network one-by-one. Last column of Table \ref{tab:EER} shows which methods are not patch-based and end-to-end. Furthermore, the performances of  $\II$ and $\DD$ as independent baselines are also reported in this table, which show that each single one of them can preform as well as previous state-of-the-arts, while our final end-to-end model, AVID, outperforms all of these methods. 
\begin{table*}[t]
\vspace{-5mm}\caption{EER and AUC perforamnce metrics on UMN dataset.} \label{tab:umn}
\begin{center}\vspace{-3mm}
\begin{tabular}{lC{2cm}C{1cm}C{2cm}C{1.6cm}C{1.5cm}C{2cm}}
\hline
&\tiny{Chaotic invariant} \cite{wu2011action}&SF \cite{mahadevan2010anomaly}&Cong~\cite{cong2011sparse}& \tiny{Saligrama}~\cite{saligrama2012video}&Li~\cite{li2014anomaly} & Ours (AVID) \\
\hline\hline
EER &5.3&12.6& 2.8& 3.4&3.7&\textbf{2.6}\\
AUC& 99.4& 94.9&\textbf{99.6}&99.5& 99.5&\textbf{99.6}\\
\hline
\vspace{-20pt}
\end{tabular}
\end{center}
\end{table*}

\begin{wrapfigure}{RB}{0.5\textwidth}
\begin{center}
\vspace{-50pt}
\includegraphics[width=\linewidth]{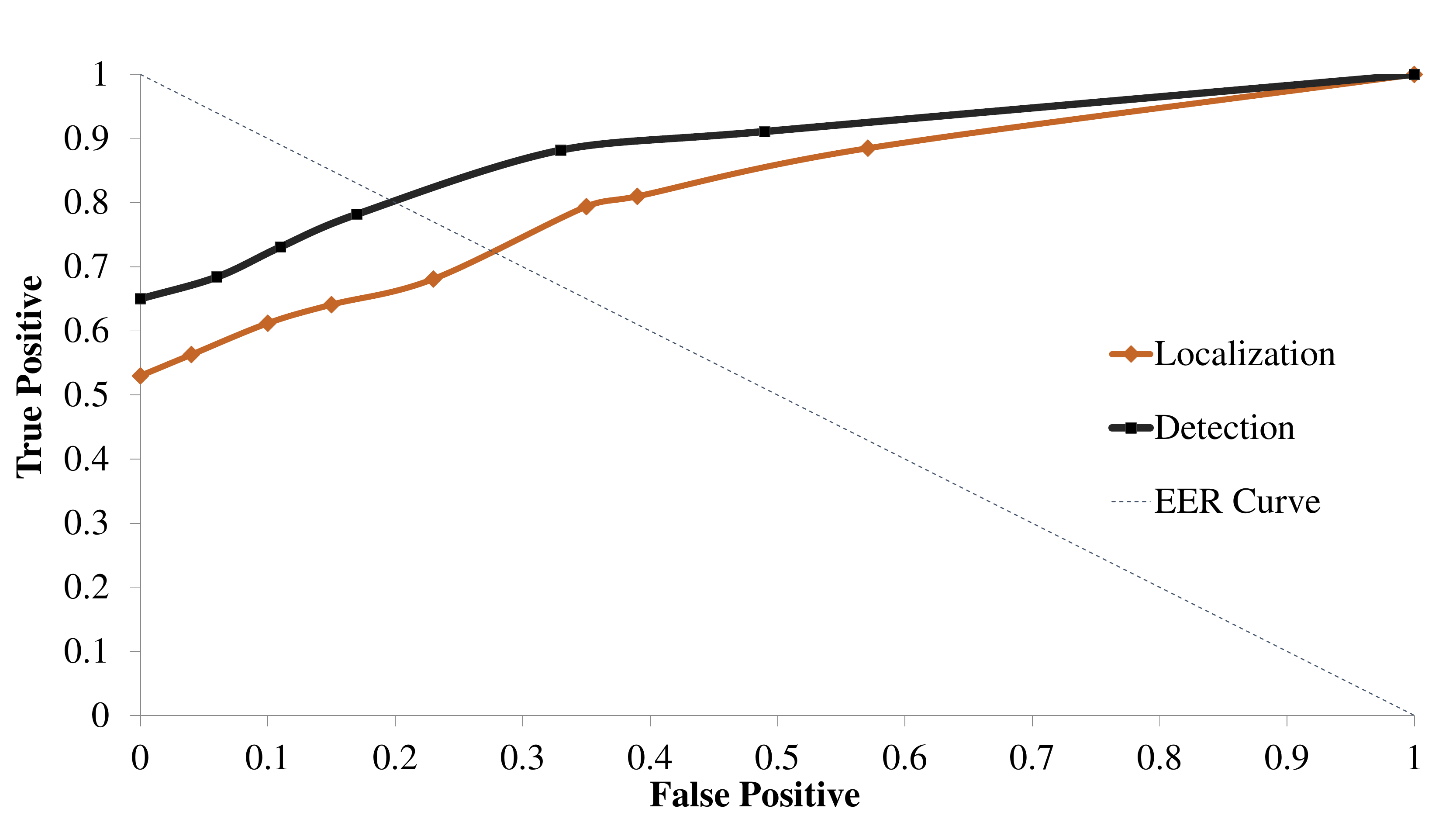}
\vspace{-20pt}
\end{center}
 \caption{ROC curves for detection and localization performance on IR-MNIST.}
 \vspace{-20pt}
 \label{fig:IR-ROC}
\end{wrapfigure}
\begin{figure}[t]
\begin{center}
\includegraphics[width=11cm, height=4cm]{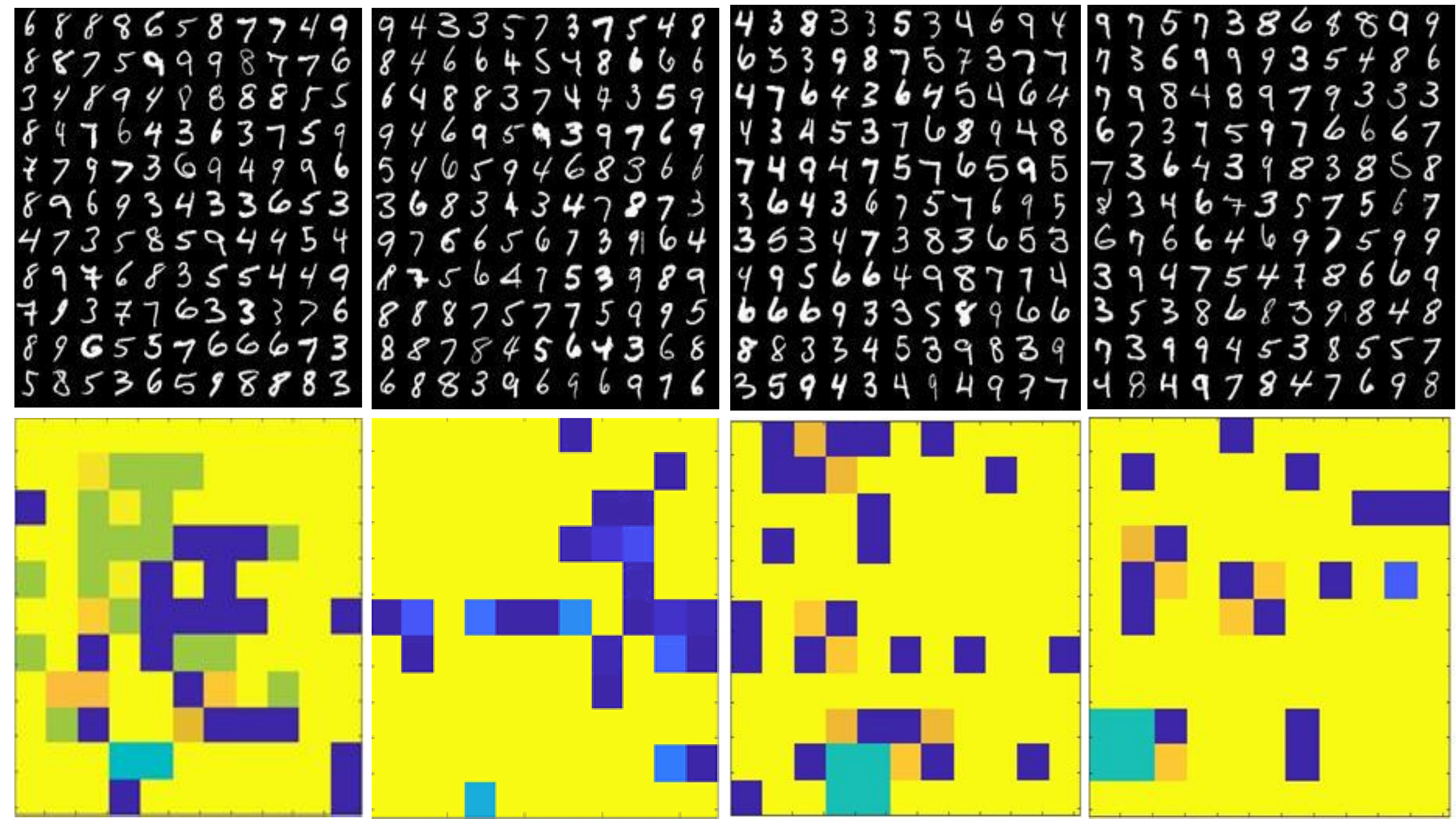}
\end{center}\vspace{-4mm}
 \caption{Examples outputs of $\DD$ on IR-MNIST. Bottom row shows (resized version of) the heat-map, for input testing images (in the top row). Here, `3' is intended as an irregular concept. 
 }
 \label{fig:D_O}
\end{figure}
\noindent\textbf{Results on UMN.} Table \ref{tab:umn} shows the irregularity detection performance in terms of equal error rate (EER) and area under the ROC curve (AUC). As discussed eariler, this dataset has some limitations, as there are only three types of anomaly scenes in the dataset with very high temporal-spatial abrupt changes between normal and abnormal frames. Also, there are no pixel-level ground truth for this dataset. Based on these limitations, to evaluate the method, EER and AUC are reported in frame-level settings. Since this dataset is simple, and irregularity localization is not important, only the global detector is used to evaluate the results. Because of the simplicity of this dataset, previous methods already performed reasonably good on this dataset. AUC of our method is comparable with the other best result, and EER of our approach is better (by 0.2 percent) than the second best.

\noindent\textbf{Results on IR-MNIST as a toy data-set.} Fig. \ref{fig:ir-mnist} confirms that the network $\II$ can properly substitute irregularity regions with a (closest) normal concept. Since `3' is considered as an irregular concept, $\II$ converted it to another digit, which is most similar to `3'. Several samples of irregular regions in Fig. \ref{fig:ir-mnist} are marked (on both the original and inpainted version of same samples). Similarly, we evaluate $\DD$ in detecting irregular regions on an image. Fig. \ref{fig:D_O} shows the heat-map of $\DD$'s output for several samples, where blue is 1 and yellow indicates 0, and other colors are in between (in a parula colormap). Note that the output is resized to have same size as the original images.
Fig. \ref{fig:IR-ROC} through the localization and detection performance on IR-MNIST dataset using the receiver operating characteristic (ROC) curve. This curve is drawn by repeatedly changing the two thresholds in Eq. \eqref{eq:2} and recording the results. Detection is just based on that if a frame contains an irregular concept (`3' digits) or not (checked over 1000 different testing samples). For localization all $11\times11$ regions on an images is considered, and if the region is correctly detected, it is counted as a true localization. So, $11\times11\times1000$ regions are evaluated. The EERs of the detection and localization are equal to $21\%$ and $29\%$, receptively. 

\subsection{Discussion} \label{sec:discussion}
\textbf{Added noise to the input of $\II $ in training phase.} 
In some cases similar to de-nosing auto-encoder \cite{vincent2008extracting}, de-noisng CNN \cite{divakar2017image} or one-class classification tasks \cite{sabokrou2018adversarially}, researchers added noise to the training data to make their network robust against noise. We also contaminated our training data with a statistical noise, with $\gamma$ indicating its intensity. This hyperparameter actually plays a very interesting role for training the model. Using this hyperparameter, we can control the learning pace between $\II$ and $\DD$. Since, $\II$ sees only normal samples during training, in the noise-free case, it can easily reconstruct them so that $\DD$ is fooled. The added noise actually makes $\II$ to learn how to inpaint and remove the irregularity to a pixel-level. Therefore, a very small value for $\gamma$ leads to a task, which is very easy for $\II$ and a very large value will mislead $\II$ from seeing the actual normal data distribution (\ie, $\PP_d$). Based on our experiments, $\gamma = 0.4$ leads to good results. From another point-of-view, $\gamma$ is a very good means to create a proper scheduling between $\II$ and $\DD$, which is a very interesting and recent topic on convergence of GANs \cite{liu2017approximation}.

\noindent\textbf{Stopping criteria.} In conventional GANs, the stop criteria is defined as when the model reaches a Nash Equilibrium between $G$ and $D$. However, for our case, the optimum point for $\II$ and $\DD$ is not often obtain at the same time. During learning of these two network, when they are competing with each other, different conditions may occur. At a time that $\DD$ is capable to efficiently classify between fake and real data (\ie, work as an accurate classifier on the validation data), we save its parameters, $\theta_{\DD}$. Also, when $\II$ generates samples as well as the normal class (\ie $\|X-\II(X)\|^2$ is in the minimum point), the parameters of $\II$, $\theta_{\II}$ are also saved. So, at different time spans $\theta_{\II}$ and $\theta_{\DD}$ are saved , during the training procedure. Similar to other GAN-style models, finding the optimum point for stopping adversarial training of $\II$+$\DD$ is a hard task.

\noindent\textbf{Mode collapse.} One of the major concerns in GANs is the mode collapse issue, which often occurs when the generator only learns a portion of the real-data distribution and outputs samples from a single mode (\ie ignores other modes). For our case, it is a different story as $\II$ directly sees all possible samples of the normal class and implicitly learns the manifold spanned by them. Reconstructing the training samples, instead of starting from a random latent space, is an acceptable way to contradict mode collapse issue \cite{makhzani2015adversarial}.  

\section{Conclusions}
In this paper, we proposed an efficient method for irregularity detection and localization of visual data (\ie images and videos). Two proposed deep networks, $\II$ and $\DD$ are adverserially trained in a self-supervised setting. $\II$ learns to efficiently reconstruct normal (regular) regions  and implicitly inpaint irregular ones. $\DD$ learns to score different regions of its input on how likely they are irregularities. Integrating the outputs of the pixel-level results from $\II$, and the patch-level results from $\DD$ provides a promising irregularity detection metric, as well as fine-segmentation of the irregularity in the visual scene. The results on several synthetic and real datasets confirm that the proposed adversarially learned network is capable of detecting irregularity, even when there are no irregular samples to use during training. Our method enjoys from the advantages of both pixel-level and patch-level methods, while not having their shortcomings.
\bibliographystyle{splncs}
\bibliography{refs}

\end{document}